%% file: main.tex
\documentclass[10pt,twocolumn,letterpaper]{article}

\usepackage{cvpr}              
\usepackage{xcolor}

\input{preamble}

\definecolor{cvprblue}{rgb}{0.21,0.49,0.74}
\usepackage[pagebackref,breaklinks,colorlinks,allcolors=cvprblue]{hyperref}

\def\paperID{17} 
\def\confName{CVPR}
\def\confYear{2026}

\title{GS4City: Hierarchical Semantic Gaussian Splatting via City-Model Priors}

\author{
Qilin Zhang$^{1,2,3}$\thanks{Equal contribution.}
\quad
Jinyu Zhu$^{1}$\footnotemark[1]
\quad
Olaf Wysocki$^{4}$
\quad
Benjamin Busam$^{1,2}$
\quad
Boris Jutzi$^{1,3}$\\[0.5em]
$^{1}$Technical University of Munich (TUM)
\quad
$^{2}$Munich Center for Machine Learning (MCML)\\
$^{3}$Karlsruhe Institute of Technology (KIT)
\quad
$^{4}$CV4DT, University of Cambridge
}

\begin{document}
\maketitle
\input{sec/0_abstract}    
\input{sec/1_intro}
\input{sec/2_rw}
\input{sec/3_method}
\input{sec/4_experiments}
\input{sec/5_conclusion}

\section*{Acknowledgements}
We gratefully acknowledge Han Sae Kim and Purdue University for providing the Gold Coast dataset and for their valuable support throughout this~work.
{
    \small
    \bibliographystyle{ieeenat_fullname}
    \bibliography{main}
}


\end{document}

%% file: preamble.tex
\usepackage{booktabs}
\usepackage{multirow}



%% file: sec/0_abstract.tex
\begin{abstract}
Recent semantic 3D Gaussian Splatting (3DGS) methods primarily rely on 2D foundation models, often yielding ambiguous boundaries and limited support for structured urban semantics. 
While city models such as CityGML encode hierarchically organized semantics together with building geometry, these labels cannot be directly mapped to Gaussian primitives. 
We present GS4City, a hierarchical semantic Gaussian Splatting method that incorporates city-model priors for urban scene understanding. 
GS4City derives reliable image-aligned masks from Level of Detail (LoD) 3 CityGML models via two-pass raycasting, explicitly using parent-child relations to validate and recover fine-grained facade elements. 
It then fuses these geometry-grounded masks with foundation-model predictions to establish scene-consistent instance correspondences, and learns a compact identity encoding for each Gaussian under joint 2D identity supervision and 3D spatial regularization.
Experiments on the TUM2TWIN and Gold Coast datasets show that GS4City effectively incorporates structured building semantics into Gaussian scene representations, outperforming existing 2D-driven semantic 3DGS baselines, including LangSplat and Gaga, by up to 15.8 IoU points in coarse building segmentation and 14.2 mIoU points in fine-grained semantic segmentation. 
By bridging structured city models and photorealistic Gaussian scene representations, GS4City enables semantically queryable and structure-aware urban reconstruction. 
Code is available at \url{https://github.com/Jinyzzz/GS4City}.
\end{abstract}

%% file: sec/1_intro.tex
\section{Introduction}
\label{sec:intro}
\begin{figure}[ht!]
    \centering
    \includegraphics[width=\linewidth]{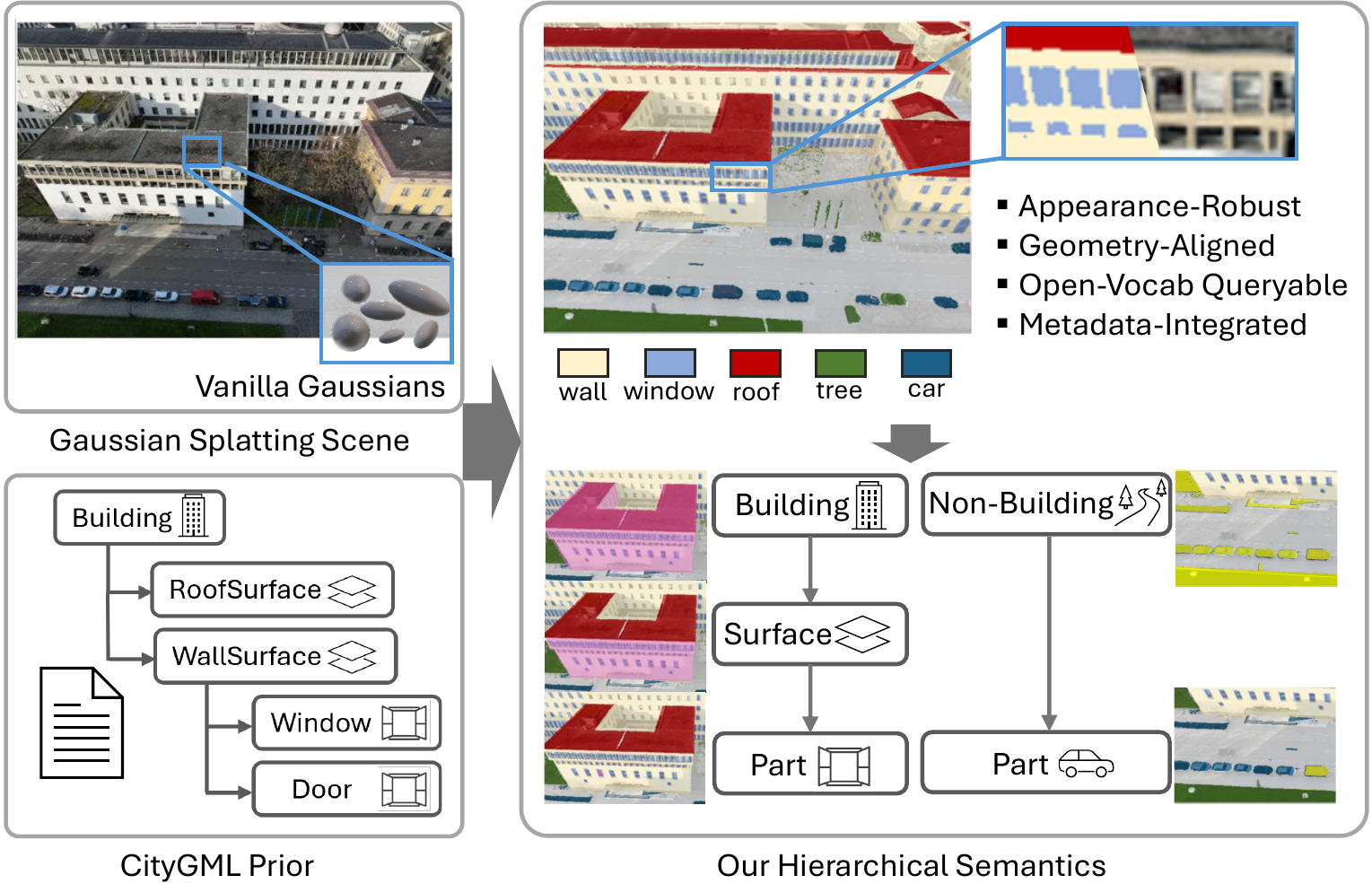}
    \caption{GS4City augments a photorealistic Gaussian scene with structured CityGML priors to produce hierarchical urban understanding. It enables appearance-robust segmentation, geometry-aligned boundaries, open-vocabulary querying, and city-model metadata integration within a unified 3DGS scene.}
    \label{fig:teaser}
\end{figure}

3D Gaussian Splatting (3DGS) \cite{kerbl20233d} has quickly evolved from real-time novel-view synthesis to a practical representation for large-scale scene reconstruction and rendering \cite{zhou2024hugs, lin2024vastgaussian, liu2024citygaussian, yan2024street, chen2024survey}. 
Recent work has further extended Gaussian scene representations beyond appearance modeling toward semantic segmentation, grouping, and open-vocabulary querying \cite{ye2024gaussian, lyu2024gaga, qin2024langsplat, wu2024opengaussian, zhou2024feature, shi2024language}. 
For urban environments, however, a key challenge remains: how to equip photorealistic Gaussian scenes with structured urban semantics that are queryable, geometrically grounded, and hierarchically organized. 
Most current semantic 3DGS methods still rely primarily on 2D foundation models \cite{qin2024langsplat, wu2024opengaussian, radford2021learning, kirillov2023segment, liu2023groundingdino}, which are effective for generic object discovery but less reliable when facade elements exhibit repeated patterns, reflective materials, cluttered foregrounds, or strong occlusions \cite{wysocki2023scan2lod3}. 
As a result, semantic boundaries often become ambiguous, and building-level part-whole structure is difficult to preserve consistently \cite{zhang2024occfaccade}. 
Yet buildings are the primary structured objects in urban environments, where urban understanding requires semantics that are both boundary-accurate and hierarchically organized across levels such as windows, doors, walls, roofs, and whole buildings \cite{biljecki2014formalisation, biljecki2017level, wysocki2023scan2lod3, tang2025texture2lod3}.

Structured 3D city models such as CityGML \cite{groger2012ogc, ledoux2019cityjson} provide a complementary source of supervision. 
They encode explicit object identities, part-whole relations, and multi-level geometric detail that are not readily available from image-based supervision \cite{biljecki2015applications}. 
However, these semantics are attached to explicit polygonal surfaces, whereas 3DGS represents scenes with alpha-composited Gaussian primitives. 
This mismatch makes direct label transfer unreliable and limits the direct use of city-model semantics for Gaussian scene understanding. 

As illustrated in \Cref{fig:teaser}, GS4City addresses this problem by using city models as structured supervision in image space and then lifting the resulting identities into the Gaussian field.
The method combines two-pass raycasting, multi-source mask fusion, and per-Gaussian identity learning to connect building entities, image masks, and Gaussian primitives within a scene-consistent semantic representation. 
This design preserves the building hierarchy while remaining compatible with open-vocabulary querying and interactive exploration of urban scenes. 
We evaluate GS4City on TUM2TWIN \cite{wysocki2026tum2twin} and Gold Coast LoD3 dataset \cite{kim2025goldcoastlod3}.
Across both datasets, GS4City improves coarse building segmentation and fine-grained semantic segmentation over existing 2D-driven semantic 3DGS baselines, including LangSplat \cite{qin2024langsplat} and Gaga \cite{lyu2024gaga}, showing that structured city-model priors can effectively extend Gaussian scenes from photorealistic rendering toward hierarchical urban understanding. 
Our main contributions are as follows:
\begin{itemize}
    \item We propose GS4City, a semantic Gaussian Splatting method that incorporates structured city-model priors for hierarchical urban understanding.
    \item We introduce a robust supervision pipeline based on two-pass raycasting and multi-source mask fusion, which integrates structured city-model semantics into reliable image-aligned supervision and establishes scene-consistent instance correspondences.
    \item We formulate semantic lifting through a compact Gaussian identity encoding with joint 2D supervision and 3D spatial regularization, enabling accurate coarse-level building segmentation, fine-grained semantic segmentation, and open-vocabulary querying in urban scenes.
\end{itemize}


%% file: sec/2_rw.tex
\section{Related Works}
\label{sec:rw}
This section reviews recent progress in semantic scene understanding for 3D Gaussian Splatting (3DGS) and the use of structured urban priors in city-scale modeling. 
\paragraph{Semantic Scene Understanding in 3DGS.}
Recent research has extended 3DGS \cite{kerbl20233d} from photorealistic rendering to semantic scene understanding \cite{piekenbrinck2025opensplat3d, fei20243d, li2025semanticsplat}, mainly along two directions: feature distillation and mask lifting \cite{he2025survey}. 
One line of work distills pretrained 2D features (e.g., CLIP \cite{radford2021learning}) into Gaussian representations \cite{zhou2024feature, qin2024langsplat, shi2024language, wu2024opengaussian}. 
Although effective for open-vocabulary querying, these continuous feature fields make it difficult to extract explicit object identities and precise geometric boundaries. 
Another line of work focuses on mask lifting, where 2D segmentation masks are associated with Gaussian primitives for sharper object-level separation \cite{ye2024gaussian, lyu2024gaga, cen2025segment, chacko2025lifting}. 
However, both directions mainly recover category- or instance-level semantics from 2D observations. 
They do not explicitly model the hierarchical part-whole relations required in structured urban environments \cite{biljecki2017level, kim2024garfield}. 
This limitation is particularly restrictive for applications such as urban digital twins and city-scale asset management, where semantics are required to be accurately aligned with geometry and consistently linked across structural levels (e.g., windows, walls, and whole buildings) \cite{biljecki2015applications}. 

\paragraph{Urban Priors for Gaussian Scene Understanding.}
In parallel, recent urban-scale 3DGS methods have shown that additional priors are essential for robust large-scale reconstruction \cite{lin2024vastgaussian, liu2024citygaussian, li2025ulsr}. 
Most of these works focus on geometric or scene-structure stabilization using LiDAR, depth, or proxy geometry, rather than structured semantic understanding \cite{yan2024street, wu2024hgs, chung2024depth, huang2025dynamic, zhang2024cdgs}. 
Standardized 3D city models, such as CityGML and CityJSON \cite{groger2012ogc, ledoux2019cityjson}, offer a compelling alternative by providing hierarchical representations of the built environment. 
While lower-level models (LoD1 and LoD2) are widely available worldwide \cite{wysocki2024reviewing} and serve as valuable, lightweight geometric priors \cite{zhang2025gs4buildings, liu2025citygo}, fine-grained urban scene understanding demands LoD3 models. 
Unlike standard meshes, LoD3 city models provide georeferenced, boundary representations (B-Rep) that explicitly encode detailed facade elements (e.g., windows and doors) alongside precise object-to-object relationships \cite{tang2025texture2lod3}. 
However, these models are typically represented as simplified polygonal meshes, lacking the photorealistic appearance and complex visual details naturally captured by 3DGS. 
Consequently, integrating the structural and semantic richness of LoD3 models with the visual fidelity of 3DGS is crucial for comprehensive urban digital twins. 
Yet, this integration poses a significant challenge due to a fundamental difference in data representation: city models attach semantics to discrete, explicit polygonal surfaces, whereas 3DGS relies on alpha-composited Gaussian primitives without explicit surface ownership. 
Bridging this gap combines the rigorous, object-level semantics of LoD3 city models with the photorealistic rendering and open-vocabulary querying capabilities of recent semantic Gaussian representations \cite{guo2026semantic, li2025langsplatv2, wu2024opengaussian, piekenbrinck2025opensplat3d}.

%% file: sec/3_method.tex
\section{Method}
\label{sec:method}
\begin{figure*}[ht!]
    \centering
    \includegraphics[width=\textwidth]{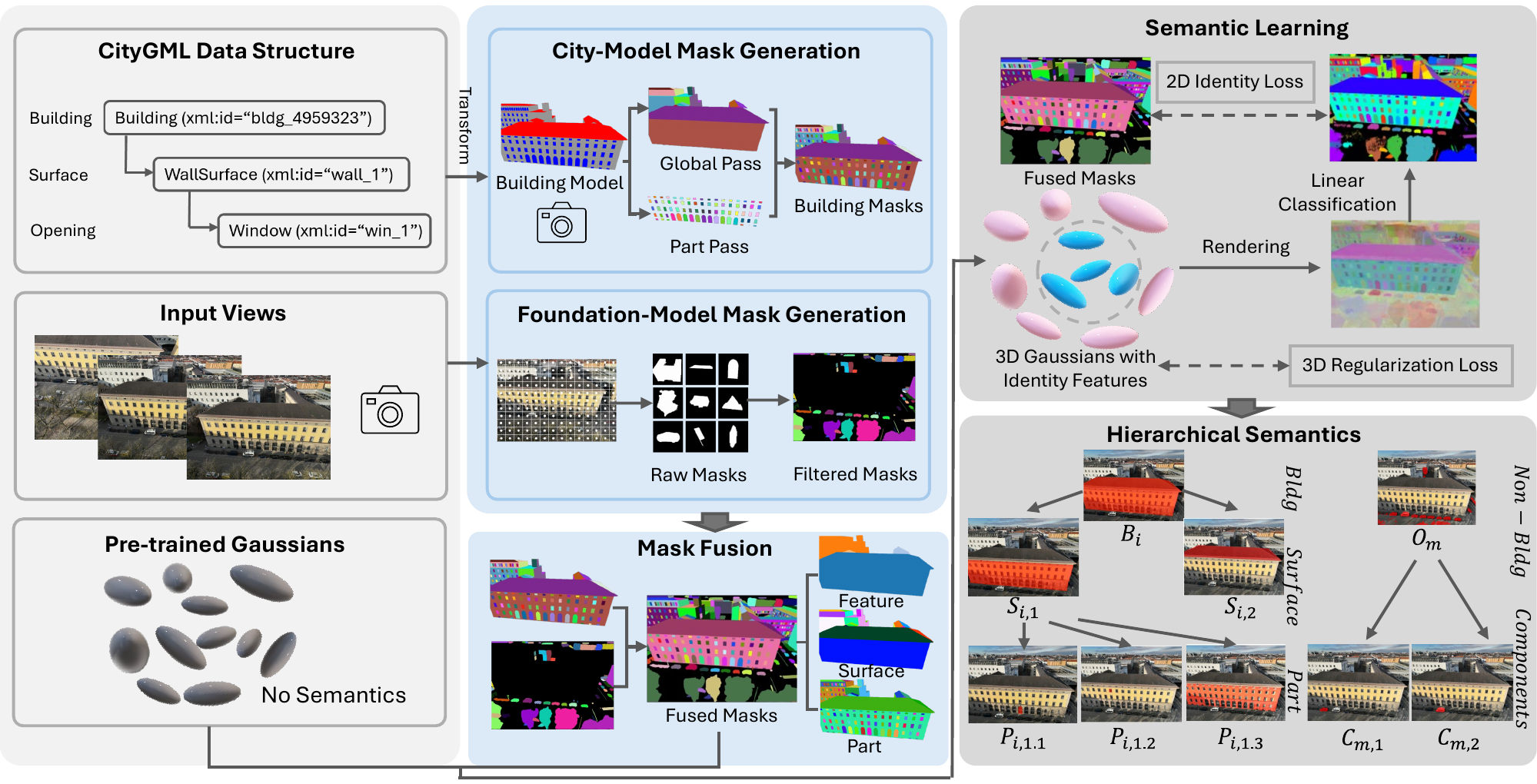}
    \caption{Overview of GS4City. Starting from an aligned LoD3 city model, calibrated images, and a pre-trained 3DGS scene, the framework first derives hierarchical city-model masks by Two-Pass Raycasting (\Cref{sec:method1}). It then generates and filters foundation-model masks, establishes scene-consistent correspondences between building entities, image masks, and Gaussian primitives through Mask Fusion and Instance Linking (\Cref{sec:method2}), and finally learns Gaussian Identity Encoding (\Cref{sec:method3}) for multi-level building parsing, semantic querying, and interactive urban-scene visualization.}
    \label{fig:overview}
\end{figure*}

Given a pre-trained 3DGS scene, a calibrated image set, and an aligned city model, GS4City learns a semantic Gaussian representation that preserves both photorealistic rendering and hierarchical urban semantics. 
Directly transferring 3D labels from city-model polygons to Gaussian primitives is unreliable. 
City models define semantics on explicit surfaces, whereas 3DGS relies on alpha-composed Gaussian primitives that lack strict surface ownership. 
In urban scenes, floaters and occlusions further aggravate this mismatch, making 3D assignment highly unstable near building boundaries. 
Therefore, GS4City constructs reliable supervision in the 2D image space. It extracts geometry-grounded masks, fuses them with full-scene foundation-model predictions, and then lifts the resulting identities into the Gaussian field. An overview is shown in \Cref{fig:overview}. 

Let $\mathcal{I}=\{I_t\}_{t=1}^{T}$ denote the training images with camera intrinsics $\mathbf{K}_t$ and extrinsics $\mathbf{E}_t$, and let $\mathcal{M}$ denote the aligned LoD3 city model. 
Following the CityGML standard, we maintain hierarchical identities for each visible entity in $\mathcal{M}$ across three levels: the \textit{feature} level for whole buildings, the \textit{surface} level for boundaries like \emph{WallSurface} and \emph{RoofSurface}, and the \textit{part} level for facade elements like \emph{Window} and \emph{Door}.  
These identities are linked by explicit parent-child relations, meaning a window belongs to a wall surface, which in turn belongs to a building. 
Our goal is to learn a Gaussian-level semantic encoding that predicts scene-consistent identities under arbitrary viewpoints.
\subsection{City-Model Semantic Extraction}
\label{sec:method1}
This subsection derives reliable image-space supervision from the aligned LoD3 city model while preserving the original building hierarchy and semantic attributes.

\paragraph{Preprocessing and Hierarchy Encoding.}
CityGML stores semantics on georeferenced polygonal building elements, whereas our supervision pipeline requires a camera-aligned triangular mesh with explicit per-face identities. 
Preprocessing performs three operations: triangulating polygonal LoD3 surfaces, aligning the mesh to the reconstruction frame, and flattening the original XML hierarchy into a semantic table. 
Each CityGML entity is assigned an integer instance ID for rendering and mask generation, while the original \texttt{objectId}, parent relation, semantic class, and attributes are preserved in the table. 
As a result, building metadata such as height and usage remains accessible through the same instance index used later for training and querying. 
Each triangle face then inherits the integer IDs of the CityGML entities it belongs to:
\begin{equation}
\ell_f = \big(id_f^{\mathrm{feat}}, id_f^{\mathrm{surf}}, id_f^{\mathrm{part}}\big),
\end{equation}
where missing levels are set to $-1$. 
Thus, a part-level face remains linked to its parent surface and building through the same ID chain for mask supervision and semantic querying.

\paragraph{Two-Pass Raycasting.}
A city-model mask is obtained by raycasting the preprocessed mesh into each training view. 
For pixel $u$ in view $t$, let $\mathcal{F}_t(u)$ be the set of faces hit by the camera ray. 
Single-pass raycasting selects the front-most face
\begin{equation}
f_t^{*}(u)=\arg\min_{f \in \mathcal{F}_t(u)} d_t(u,f),
\end{equation}
where $d_t(u,f)$ is the intersection depth, and the corresponding hierarchical label tuple is
\begin{equation}
M_t^{\mathrm{global}}(u)=\ell_{f_t^{*}(u)}.
\end{equation}
This provides feature-, surface-, and part-level labels in a single pass. 
However, single-pass projection often fails around facade openings. 
If triangulated wall faces cover a window or door region, the front-most hit is assigned to the wall, and the projected label loses its correct part identity. 
To recover them, we explicitly leverage CityGML parent-child relations. 
A second pass is applied to part-level geometries only, producing a part depth map $\widetilde{D}_t$, candidate part IDs $\tilde{p}_t(u)$, and their parent-surface IDs $\tilde{s}_t(u)$. 
Let $D_t(u)$ denote the depth of the visible face selected in the global pass, and let $s_t(u)$ be its surface ID. 
The part-pass result is accepted only when
\begin{equation}
\gamma_t(u)=
\big(\tilde{s}_t(u)=s_t(u)\big)
\land
\big(\widetilde{D}_t(u)-D_t(u)\le\tau\big),
\end{equation}
where $\tau$ is a depth tolerance. 
The recovered part label is then
\begin{equation}
p_t(u)=
\begin{cases}
\tilde{p}_t(u), & \gamma_t(u),\\
-1, & \text{otherwise}.
\end{cases}
\end{equation}
The raycasting stage produces feature-, surface-, and part-level ID maps. 
For subsequent fusion and learning, a single city-model instance map is constructed from these hierarchical labels by selecting the finest valid level: 
\begin{equation}
M_t^{\mathrm{city}}(u)=
\begin{cases}
p_t(u), & p_t(u)\neq -1,\\
id_t^{\mathrm{surf}}(u), & p_t(u)=-1 \land id_t^{\mathrm{surf}}(u)\neq -1,\\
id_t^{\mathrm{feat}}(u), & \text{otherwise}.
\end{cases}
\end{equation}
Thus, part IDs override surface IDs, and surface IDs override feature IDs. 
The parent-child hierarchy remains available through the semantic table. 
This two-pass construction yields city-model masks with substantially fewer missing facade elements than single-pass projection.

\subsection{Mask Fusion and Instance Linking}
\label{sec:method2}
City models provide precise supervision for buildings but do not cover the full scene. 
This subsection complements $M_t^{\mathrm{city}}$ with image-based masks, establishes cross-view instance IDs, and forms the full-scene supervision used later for Gaussian identity learning.

\paragraph{Foundation-model Mask Filtering.}
\begin{figure}[ht!]
    \centering
    \includegraphics[width=\linewidth]{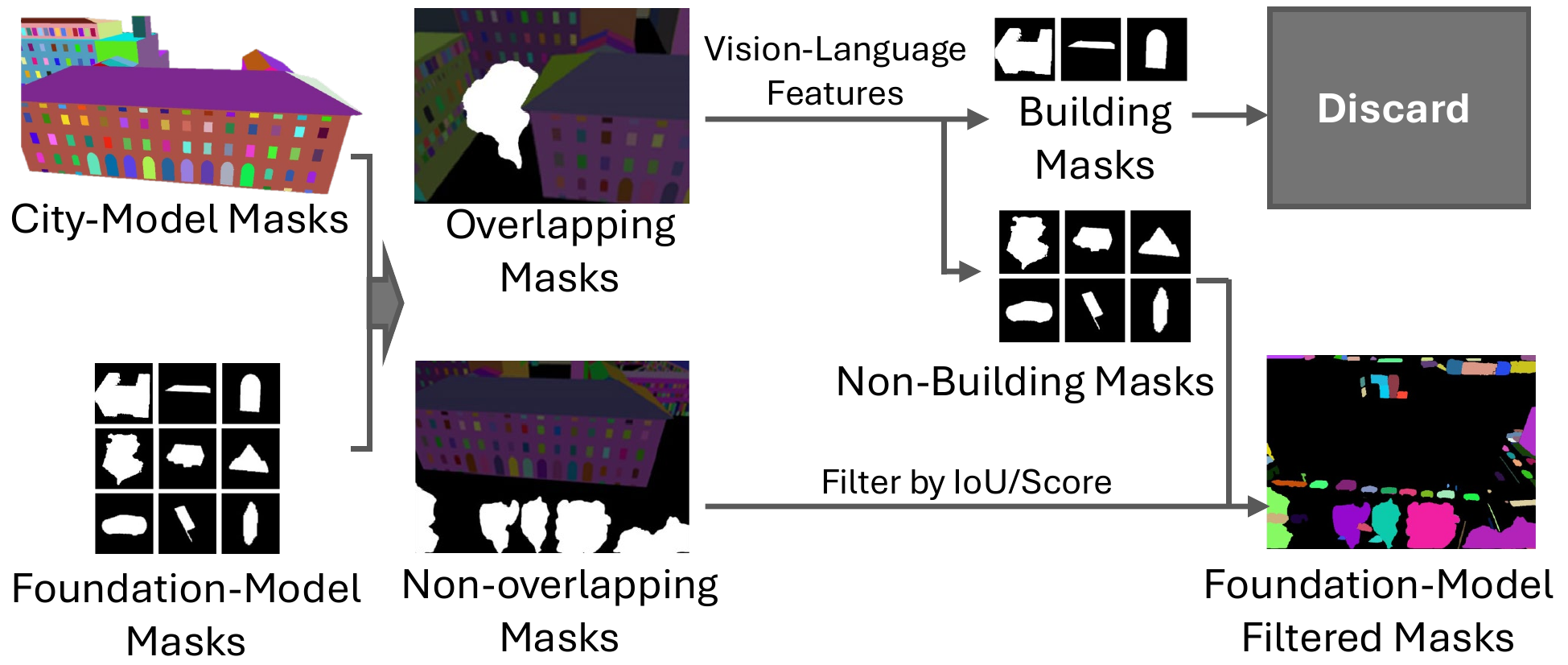}
    \caption{Foundation-model mask filtering. Raw image-based masks are compared with the city-model building-support region. High-overlap masks are disambiguated by a vision-language model into building-related versus foreground candidates, yielding cleaned image-based masks for cross-view association.}
    \label{fig:mask_filtering}
\end{figure}
For each view $t$, an image-based foundation model \cite{kirillov2023segment} produces a set of raw masks $\mathcal{S}_t=\{s_i\}$. 
The filtering workflow is illustrated in \Cref{fig:mask_filtering}. 
An initial filtering step removes masks with low model-predicted quality scores or insufficient pixel support:
\begin{equation}
\mathcal{S}_t^{\mathrm{qc}}=
\left\{
s_i \in \mathcal{S}_t
\ \middle|\
q(s_i)\ge\tau_q,\ |s_i|\ge\tau_a
\right\},
\end{equation}
where $q(s_i)$ denotes the quality score predicted by the mask generator and $|s_i|$ is the mask area in pixels. 
From the city-model supervision, the building-support region is defined as
\begin{equation}
\Omega_t^{\mathrm{feat}}=\{u \mid id_t^{\mathrm{feat}}(u)\neq -1\}.
\end{equation}
Each mask $s_i \in \mathcal{S}_t^{\mathrm{qc}}$ is then compared with this region:
\begin{equation}
R_{\mathrm{ov}}(s_i)=\frac{|s_i \cap \Omega_t^{\mathrm{feat}}|}{|s_i|}.
\end{equation} 
Masks with small overlap are kept as non-building candidates. 
Large-overlap masks, which may represent either buildings or foreground objects, are disambiguated using a vision-language model \cite{radford2021learning}. 
Let $\mathbf{f}_{\mathrm{img}}(s_i)$ be the image feature of the cropped mask, $\mathcal{T}_{\mathrm{city}}$ and $\mathcal{T}_{\mathrm{fore}}$ denote building-related and foreground prompt sets.
The two scores are
\begin{equation}
S_{\mathrm{city}}=\max_{q \in \mathcal{T}_{\mathrm{city}}}
\cos\!\big(\mathbf{f}_{\mathrm{img}}(s_i), \mathbf{f}_{q}\big),
\end{equation}
\begin{equation}
S_{\mathrm{fore}}=\max_{q \in \mathcal{T}_{\mathrm{fore}}}
\cos\!\big(\mathbf{f}_{\mathrm{img}}(s_i), \mathbf{f}_{q}\big),
\end{equation}
where $\mathbf{f}_{q}$ denotes the text feature of prompt $q$.  
The mask is kept only if
\begin{equation}
S_{\mathrm{fore}} > S_{\mathrm{city}} + \delta,
\end{equation}
where $\delta$ is a margin. 
This yields a cleaned mask set $\widehat{\mathcal{S}}_t$ that mainly captures non-building foreground objects.

\paragraph{Cross-view Association.}
The cleaned masks $\widehat{\mathcal{S}}_t$ are still view-specific. 
To assign scene-consistent instance IDs, the pre-trained Gaussian scene is used only as a geometric bridge, not yet as a semantic model. 
Inspired by the 3D-aware Gaussian grouping paradigm of Gaga \cite{lyu2024gaga}, cross-view association is performed by combining geometric overlap with vision-language semantic verification. 
For each mask $s_i \in \widehat{\mathcal{S}}_t$, the Gaussian centers $\mu_j$ are projected into the image by
\begin{equation}
\mathbf{p}_{t,j}=\Pi(\mathbf{K}_t,\mathbf{E}_t,\mu_j),
\end{equation}
and only the points inside $s_i$ that pass depth-based foreground filtering are retained. 
This defines a candidate Gaussian set $\mathcal{C}_t(s_i)$. 
The candidate is matched against the current global groups $\{\mathcal{G}_k\}$ by
\begin{equation}
R_{\mathrm{geo}}(s_i,\mathcal{G}_k)=
\frac{|\mathcal{C}_t(s_i)\cap \mathcal{G}_k|}{|\mathcal{C}_t(s_i)|}.
\end{equation}
Let
\begin{equation}
k^{*}=\arg\max_k R_{\mathrm{geo}}(s_i,\mathcal{G}_k).
\end{equation}
Mask $s_i$ is merged into $\mathcal{G}_{k^{*}}$ only when the geometric overlap is high and its vision-language feature is semantically close to the aggregated feature of that group:
\begin{equation}
R_{\mathrm{geo}}(s_i,\mathcal{G}_{k^{*}})>\tau_{\mathrm{geo}}
\ \land\
\cos\!\big(\mathbf{f}_{\mathrm{img}}(s_i),\bar{\mathbf{f}}_{k^{*}}\big)>\tau_{\mathrm{sim}},
\end{equation}
where $\bar{\mathbf{f}}_{k^{*}}$ is the running average vision-language feature of group $\mathcal{G}_{k^{*}}$. 
Otherwise, a new group is initialized. 
After all views are processed, groups with insufficient multi-view support are removed and far-field groups are filtered by their distance to the camera set. 
The remaining group IDs are then transferred back to each view, yielding stable image-based masks $M_t^{\mathrm{img}}$ with scene-consistent instance IDs.

\paragraph{Mask Fusion and Feature Aggregation.}
The City-model mask $M_t^{\mathrm{city}}$ and the associated image-based mask $M_t^{\mathrm{img}}$ are merged into a unified label map $M_t$. 
Since the retained image-based masks correspond either to non-building regions or to valid foreground occluders, an image-mask-priority fusion rule is used:
\begin{equation}
\widehat{M}_t(u)=
\begin{cases}
M_t^{\mathrm{img}}(u)+\Delta, & M_t^{\mathrm{img}}(u)>0,\\
0, & \text{otherwise},
\end{cases}
\end{equation}
\begin{equation}
M_t(u)=
\begin{cases}
\widehat{M}_t(u), & \widehat{M}_t(u)>0,\\
M_t^{\mathrm{city}}(u), & M_t^{\mathrm{city}}(u)>0,\\
0, & \text{otherwise}.
\end{cases}
\end{equation}
Here $\Delta$ is a fixed ID offset that keeps image-based instances disjoint from city-model IDs. 
This fusion preserves accurate building-part supervision while extending the label space to the full scene. 

To support open-vocabulary querying, a vision-language feature is associated with each fused instance ID \cite{radford2021learning}. 
For image-based groups, the aggregated feature $\bar{\mathbf{f}}_{k}$ is updated during association. 
For city-model instances, image features are extracted from all visible masked crops and averaged across views:
\begin{equation}
\bar{\mathbf{f}}_{k}=
\mathrm{Normalize}\!\left(
\frac{1}{N_k}\sum_{j=1}^{N_k}\mathbf{f}_{k,j}
\right),
\end{equation}
where $\mathbf{f}_{k,j}$ is the image feature of instance $k$ in view $j$. 
The final representation therefore establishes a scene-consistent identity chain linking city-model entities, image masks, Gaussian groups, and per-instance semantic features. 

\subsection{Gaussian Identity Learning}
\label{sec:method3}
Starting from the fused supervision map $M_t$, this stage lifts scene-consistent instance labels into the pre-trained Gaussian scene.

\paragraph{Identity Encoding.}
Geometry and appearance parameters of the pre-trained 3DGS are kept fixed, and each Gaussian $g_j$ is augmented with a learnable identity code $\mathbf{e}_j \in \mathbb{R}^{D}$. 
These codes are view-independent and rendered with the same alpha-compositing rule used for color rendering:
\begin{equation}
\mathbf{E}(u)=\sum_{j \in \mathcal{R}(u)}
\mathbf{e}_j \, \alpha'_j
\prod_{k<j}(1-\alpha'_k),
\end{equation}
where $\mathcal{R}(u)$ is the depth-ordered set of Gaussians contributing to pixel $u$ and $\alpha'_j$ is the projected opacity contribution of Gaussian $j$. 
The rendered feature $\mathbf{E}(u)$ is mapped by a linear classifier $f(\cdot)$ into the scene-consistent identity space defined by the fused label map $M_t$.

\paragraph{2D Identity Supervision.}
The primary supervision is the fused label map produced in the previous subsection. 
For view $t$, the rendered identity logits are compared against $M_t$ with a chunked cross-entropy loss:
\begin{equation}
\mathcal{L}_{2\mathrm{D}}=
\frac{1}{|\Omega_t|}
\sum_{u \in \Omega_t}
\mathrm{CE}\!\left(f(\mathbf{E}(u)), M_t(u)\right),
\end{equation}
where $\Omega_t$ is the valid rendered pixel set. 

\paragraph{3D Spatial Regularization.}
Purely 2D supervision leaves weakly observed Gaussians underconstrained. 
Local semantic consistency is therefore imposed in 3D by sampling $m$ Gaussians, retrieving $k$ nearest neighbors, and penalizing disagreement between predicted class distributions:
\begin{equation}
\mathbf{p}_j=\mathrm{softmax}\!\big(f(\mathbf{e}_j)\big),
\end{equation}
\begin{equation}
\mathcal{L}_{3\mathrm{D}}=
\frac{1}{m k K}
\sum_{j=1}^{m}
\sum_{i=1}^{k}
\mathrm{KL}\!\left(\mathbf{p}_j \,\|\, \mathbf{p}_{j_i}\right),
\end{equation}
where $K$ is the number of labels including background, and $\mathbf{p}_{j_i}$ is the predicted distribution of the $i$-th neighbor of Gaussian $j$. 
The total objective is
\begin{equation}
\mathcal{L}=
\mathcal{L}_{2\mathrm{D}}+
\lambda_{3\mathrm{D}}\,\rho_t\,\mathcal{L}_{3\mathrm{D}},
\end{equation}
where $\rho_t \in \{0,1\}$ activates the 3D term only at scheduled iterations. 
After training, each Gaussian and rendered pixel is assigned a scene-consistent instance ID. 
Through the identity chain constructed above, this ID can be traced back to the semantic table, making the learned Gaussian scene queryable at feature, surface, and part levels together with the associated city-model attributes.

%% file: sec/4_experiments.tex
\section{Experiments}
\label{sec:experiments}
This section introduces the datasets, baselines, evaluation protocol, and implementation settings, and then reports results on building understanding and prompt-driven querying across two urban datasets. 

\subsection{Experimental Setup}
\paragraph{Datasets and Evaluation.}
Experiments are conducted on two urban LoD3 benchmarks: TUM2TWIN \cite{wysocki2026tum2twin} and Gold Coast LoD3 \cite{kim2025goldcoastlod3}. 
TUM2TWIN provides georeferenced UAV imagery, aligned LoD3 CityGML models, and building-related ground truth from ZAHA \cite{wysocki2025zaha}. 
Gold Coast provides oblique UAV imagery, LoD3 building models, and a labeled point cloud, whose projected labels are used to obtain building-part ground-truth masks. 
For non-building categories on both datasets, dense manual annotations are unavailable, so auxiliary 2D pseudo labels are used for the same set of foreground categories across all methods. 
All methods are evaluated under a unified prompt-driven protocol, in which each text query yields a binary prediction mask that is compared against the corresponding ground truth. 
GS4City uses a hybrid querying scheme that combines semantic labels with open-vocabulary matching, whereas LangSplat uses learned scene features for open-vocabulary querying. 
Gaga uses GroundingDINO \cite{liu2023groundingdino} to localize the queried object and select the associated instance mask. 
Evaluation is reported at two semantic levels. The \emph{coarse level} groups the scene into Building and Non-Building and is reported with IoU, accuracy, precision, and recall. The \emph{fine-grained level} resolves wall, roof, window, door, tree, person, and car, and is reported with mIoU together with per-class IoU and precision. 

\paragraph{Baselines and Implementation.}
As GS4City combines mask lifting with open-vocabulary querying, two representative semantic 3DGS baselines are considered: Gaga \cite{lyu2024gaga}, representing mask-lifting methods without native text querying and typically achieving more accurate segmentation, and LangSplat \cite{qin2024langsplat}, representing feature-distillation methods with native open-vocabulary retrieval. 
All methods start from pre-trained 3DGS reconstructions trained for 30k iterations and use the same train/test split on each dataset. 
In the two-pass raycasting stage, the depth tolerance is set to $\tau = 0.5$\,m to accommodate typical facade recesses.  

\subsection{Results and Discussion}
We evaluate GS4City through qualitative comparisons, quantitative results at the coarse and fine-grained semantic levels, and an ablation study of the cross-view association.

\begin{figure*}[t]
\centering
\includegraphics[width=\textwidth]{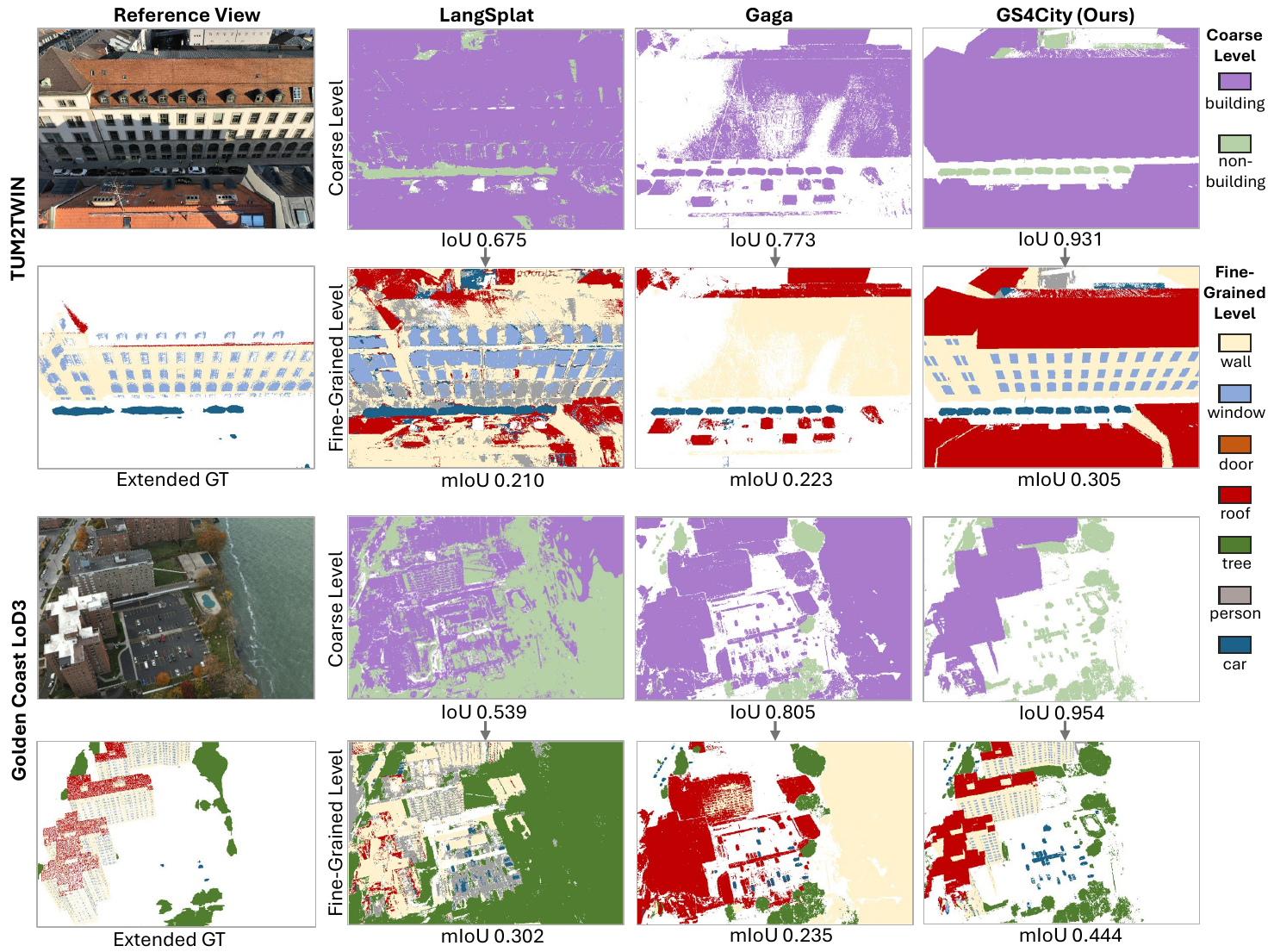}
\caption{Qualitative results on representative scenes from TUM2TWIN and Gold Coast at both the coarse and fine-grained semantic levels. The first column shows the reference RGB views together with the extended ground truth derived from labeled point clouds and auxiliary 2D pseudo labels. The remaining columns show the predictions of LangSplat, Gaga, and GS4City. GS4City provides more complete Building/Non-Building separation at the coarse level and more accurate fine-grained segmentation of challenging building parts, such as windows and roofs, while avoiding incorrect semantic assignments in non-building regions, such as water surfaces.}
\label{fig:qualitative_results}
\end{figure*}

\Cref{fig:qualitative_results} compares the three methods on representative urban scenes from both datasets. 
LangSplat tends to assign labels to nearly all visible pixels, which yields broad scene coverage but also fragmented predictions that are poorly aligned with building geometry.
It can roughly localize windows and roofs, yet the predicted regions are often noisy and spatially inconsistent. 
Gaga produces more coherent regions overall, but it still struggles to isolate difficult facade elements such as windows and often absorbs them into coarser wall or roof regions. 
This behavior is particularly visible on Gold Coast, where planar non-building areas such as water can be confused with wall surfaces. 
In contrast, GS4City yields stable coarse-level Building/Non-Building separation, recovers fine-grained building parts more accurately, and is less prone to assigning labels to uncertain non-building regions, leading to cleaner vehicle masks and fewer obvious semantic failures. 
These qualitative results support the role of city-model supervision in providing sharper geometric boundaries and more reliable structural identities.

\begin{table*}[t]
\centering
\scriptsize
\setlength{\tabcolsep}{5pt}
\caption{Overall results at the coarse and fine-grained semantic levels. Coarse-level metrics evaluate Building/Non-Building grouping, while the fine-grained part reports overall mIoU together with per-class IoU. All values are reported in \%.}
\label{tab:overall_results}
\resizebox{0.97\textwidth}{!}{%
\begin{tabular}{@{}llcccccccccccc@{}}
\toprule
\multirow{2}{*}{\textbf{Dataset}} & \multirow{2}{*}{\textbf{Method}} & \multicolumn{4}{c}{\textbf{Coarse Level}} & \multicolumn{8}{c}{\textbf{Fine-Grained Level}} \\
\cmidrule(lr){3-6} \cmidrule(l){7-14}
 &  & \textbf{IoU$\uparrow$} & \textbf{Acc.$\uparrow$} & \textbf{Prec.$\uparrow$} & \textbf{Rec.$\uparrow$} & \textbf{mIoU$\uparrow$} & \textbf{Door} & \textbf{Roof} & \textbf{Wall} & \textbf{Window} & \textbf{Car} & \textbf{Person} & \textbf{Tree} \\
\midrule
\multirow{3}{*}{TUM2TWIN}
& LangSplat & 67.5 & 71.2 & 86.2 & 75.8 & 21.0 & 0.1 & 3.3 & 39.3 & 12.9 & 61.7 & 4.5 & \textbf{30.5} \\
& Gaga & 77.3 & 80.8 & 91.2 & 80.9 & 22.3 & 0.0 & 1.1 & \textbf{59.3} & 1.2 & 78.6 & 6.1 & 18.3 \\
& GS4City (Ours) & \textbf{93.1} & \textbf{94.6} & \textbf{96.2} & \textbf{96.6} & \textbf{30.5} & \textbf{3.6} & \textbf{19.8} & 59.2 & \textbf{25.6} & \textbf{82.3} & \textbf{9.0} & 18.8 \\
\midrule
\multirow{3}{*}{Gold Coast}
& LangSplat & 53.9 & 71.6 & 86.7 & 59.5 & 30.2 & 0.0 & 5.0 & 36.4 & 0.0 & 27.3 & -- & \textbf{83.8} \\
& Gaga & 80.5 & 87.7 & 87.6 & 91.1 & 23.5 & 0.0 & 20.6 & 2.2 & 0.0 & 33.3 & -- & 72.9 \\
& GS4City (Ours) & \textbf{95.4} & \textbf{97.1} & \textbf{99.9} & \textbf{95.5} & \textbf{44.4} & -- & \textbf{48.6} & \textbf{48.7} & \textbf{19.5} & \textbf{61.2} & -- & 57.4 \\
\bottomrule
\end{tabular}%
}
\end{table*}

\Cref{tab:overall_results} summarizes the quantitative results at both semantic levels. 
At the coarse level, GS4City achieves the strongest Building/Non-Building separation on both datasets, reaching 93.1\% IoU on TUM2TWIN and 95.4\% IoU on Gold Coast. 
It also yields the best accuracy, precision, and recall in both cases. 
Together, these results indicate that GS4City can separate building and non-building regions both more completely and more reliably, with fewer obvious over-segmentation errors and fewer missed building regions than the competing methods. 
At the fine-grained level, GS4City again obtains the best overall mIoU, improving from 22.3\% and 21.0\% to 30.5\% on TUM2TWIN and from 23.5\% and 30.2\% to 44.4\% on Gold Coast. 
Although LangSplat remains stronger on the irregular vegetation category \emph{tree}, suggesting that generic vision-language feature distillation can still be advantageous for categories that lack stable structural priors, the per-class IoU further shows that GS4City delivers the largest gains on building-related categories and other objects with regular geometry, including roof, wall, window, and car, which is consistent with the qualitative results in \Cref{fig:qualitative_results}. 
\begin{table}[t]
\centering
\small
\setlength{\tabcolsep}{1pt}
\caption{Per-class precision at the fine-grained semantic level. Classes absent from a dataset split are marked as ``--''. All values are reported in \%.}
\label{tab:finegrained_precision}
\resizebox{0.49\textwidth}{!}{%
\begin{tabular}{@{}llcccccccc@{}}
\toprule
\multirow{2}{*}{\textbf{Dataset}} & \multirow{2}{*}{\textbf{Method}} & \multicolumn{7}{c}{\textbf{Per-class Prec.$\uparrow$}} \\
\cmidrule(lr){3-9}
 &  & \textbf{Door} & \textbf{Roof} & \textbf{Wall} & \textbf{Wind.} & \textbf{Car} & \textbf{Pers.} & \textbf{Tree} \\
\midrule
\multirow{3}{*}{TUM2TWIN}
& LangSplat & 0.1 & 7.6 & 70.4 & 22.5 & 61.7 & 4.7 & 43.9 \\
& Gaga & 0.0 & 2.8 & 76.4 & 30.0 & 78.6 & 23.0 & 41.3 \\
& GS4City (Ours) & \textbf{5.4} & \textbf{21.7} & \textbf{81.6} & \textbf{39.9} & \textbf{82.3} & \textbf{39.1} & \textbf{55.0} \\
\midrule
\multirow{3}{*}{Gold Coast}
& LangSplat & -- & 23.5 & 57.1 & 0.0 & 27.3 & -- & 86.1 \\
& Gaga & -- & 23.7 & 17.3 & 3.4 & 33.3 & -- & \textbf{99.2} \\
& GS4City (Ours) & -- & \textbf{50.8} & \textbf{81.9} & \textbf{46.4} & \textbf{61.2} & -- & 98.7 \\
\bottomrule
\end{tabular}
}
\end{table}

\Cref{tab:finegrained_precision} further clarifies this trend by reporting per-class precision. 
GS4City achieves the strongest precision on most structured categories, which shows that its improvements are not only due to broader coverage but also to cleaner and more selective semantic assignments. 
The advantage is particularly clear on Gold Coast, where GS4City substantially increases precision on the main building-related categories while avoiding obvious confusion with large planar background regions. 
For \emph{tree}, where LangSplat attains the best IoU, GS4City remains competitive in precision, indicating that its weaker IoU on irregular classes is mainly associated with incomplete coverage rather than severe false positives. 

These results are consistent with the design of GS4City. 
Structured city-model supervision provides sharper boundaries and explicit part-whole relations, reducing the ambiguity caused by repeated facade patterns and visually similar building elements. 
At the same time, the scene-consistent identity chain supports more reliable instance retrieval than feature-only matching when multiple facade regions share a similar appearance. 
Overall, GS4City achieves strong performance across categories, with the largest gains on those supported by stable structural priors.

\paragraph{Ablation Study.}
\begin{table}[t]
\centering
\footnotesize
\setlength{\tabcolsep}{7pt}
\caption{Ablation on the minimal-view filter in cross-view association on a challenging TUM2TWIN subset with strong tree occlusions. LangSplat and Gaga are included as reference baselines. The reported mIoU is measured at the fine-grained semantic level.}
\label{tab:ablation_minview}
\begin{tabular}{@{}lcccc@{}}
\toprule
\textbf{Method} & \textbf{$m_{\mathrm{view}}$} & \textbf{mIoU$\uparrow$} & \textbf{Time (min)$\downarrow$} & \textbf{mIoU/min$\uparrow$} \\
\midrule
LangSplat & -- & 15.8 & \textbf{39.2} & 0.40 \\
Gaga & -- & 18.7 & 43 & 0.43 \\
\midrule
GS4City (Ours) & 1 & \textbf{25.8} & 84.7 & 0.30 \\
GS4City (Ours) & 3 & 24.0 & 55.0 & \textbf{0.44} \\
GS4City (Ours) & 5 & 21.9 & 51.8 & 0.42 \\
\bottomrule
\end{tabular}
\end{table}
The minimal-view filter in cross-view association controls a key trade-off in GS4City: lower thresholds preserve semantic coverage, whereas higher thresholds remove unstable instances and reduce computation. 
We ablate the threshold $m_{\mathrm{view}}$, which keeps only groups observed in at least $m_{\mathrm{view}}$ views, on a challenging TUM2TWIN subset with frequent tree occlusions over building facades. 
\Cref{tab:ablation_minview} reports the results for $m_{\mathrm{view}}\in\{1,3,5\}$ together with LangSplat and Gaga as reference baselines; 
the reported mIoU corresponds to the fine-grained semantic level, and time-normalized mIoU (mIoU/min) summarizes the accuracy-efficiency trade-off. 
Setting $m_{\mathrm{view}}=1$ yields the highest absolute mIoU among the GS4City variants, but also the longest runtime. 
Increasing the threshold to $m_{\mathrm{view}}=5$ reduces runtime but lowers mIoU more noticeably, indicating that aggressive filtering removes useful but sparsely observed instances in heavily occluded scenes. 
The intermediate setting $m_{\mathrm{view}}=3$ provides the best balance and is therefore used in the main experiments. Although GS4City with $m_{\mathrm{view}}=3$ is slightly slower than LangSplat and Gaga, it achieves the highest time-normalized mIoU at 0.44, compared with 0.40 and 0.43 for the two baselines, indicating the best overall accuracy-efficiency trade-off in this challenging subset. 

\paragraph{Limitations and Future Work.}
GS4City is designed for urban scenes with aligned LoD3 city models as structured supervision, which defines its current scope. 
Its performance depends on both the geometric alignment and the semantic completeness of the city model, since missing or simplified facade elements directly weaken the projected supervision. 
Future work includes more robust alignment and update schemes, and explores whether weaker priors such as LoD2 models can be combined with semantic segmentation and reasoning to provide finer-grained structured supervision. 
Another direction is to investigate how structured semantics can be leveraged to improve reconstruction completeness and geometric accuracy. 

%% file: sec/5_conclusion.tex
\section{Conclusion}
\label{sec:conclusion}
This paper presents GS4City, a hierarchical semantic Gaussian Splatting method that integrates city-model priors into urban Gaussian scenes. 
The method transfers LoD3 city-model semantics into reliable image-aligned supervision, establishes scene-consistent identities across city-model entities, image masks, and Gaussian primitives, and lifts them into the Gaussian field through identity encoding learning.
Experiments on TUM2TWIN and Gold Coast dataset show that this design improves coarse-level Building/Non-Building segmentation and fine-grained urban semantic segmentation by up to 15.8 points in IoU and 14.2 points in mIoU over existing semantic 3DGS baselines. 
The gains are strongest on structured categories such as roofs, walls, windows, and cars, while predictions on irregular categories such as trees remain selective but precise. 
Ablation results further show that GS4City maintains a favorable accuracy-efficiency trade-off, achieving the best time-normalized fine-grained mIoU with only a modest increase in runtime. 
More importantly, by preserving the hierarchy encoded in the city model, GS4City links Gaussian identities to building-, surface-, and part-level semantics together with associated city-model metadata, enabling semantic hierarchical querying and structured urban information retrieval.
In this sense, GS4City extends Gaussian scene representations from appearance-centered rendering toward hierarchical urban understanding.